\documentclass[sigconf]{acmart}

\usepackage[utf8]{inputenc} 
\usepackage[T1]{fontenc}    
\usepackage{hyperref}       
\usepackage{url}            
\usepackage{booktabs}       
\usepackage{amsfonts}       
\usepackage{nicefrac}       
\usepackage{microtype}      
\usepackage{xcolor}         
\usepackage{multirow}
\usepackage{graphicx}
\usepackage{amsmath,epsfig}
\usepackage{wrapfig}

\usepackage{booktabs}
\AtBeginDocument{%
  \providecommand\BibTeX{{%
    \normalfont B\kern-0.5em{\scshape i\kern-0.25em b}\kern-0.8em\TeX}}}

\begin{document}

\title{Self-Supervised Graph Neural Network for Multi-Source Domain Adaptation}


\author{Jin Yuan}
\affiliation{%
\institution{Southeast University}
\city{Nanjing}
\country{China}
}
\email{yuanjin@seu.edu.cn}

\author{Feng Hou}
\affiliation{%
\institution{University of Chinese Academy of Sciences}
\city{Beijing}
\country{China}
}
\email{houfeng19@mails.ucas.ac.cn}

\author{Yangzhou Du}
\affiliation{%
\institution{Lenovo Research}
\city{Beijing}
\country{China}
}
\email{duyz1@lenovo.com}

\author{Zhongchao Shi}
\affiliation{%
\institution{Lenovo Research}
\city{Beijing}
\country{China}
}
\email{shizc2@lenovo.com}

\author{Xin Geng}
\affiliation{%
\institution{Southeast University}
\city{Nanjing}
\country{China}
}
\email{xgeng@seu.edu.cn}

\author{Jianping Fan}
\affiliation{%
\institution{Lenovo Research}
\city{Beijing}
\country{China}
}
\email{jfan1@lenovo.com}

\author{Yong Rui}
\affiliation{%
\institution{Lenovo Research}
\city{Beijing}
\country{China}
}
\email{yongrui@lenovo.com}

\begin{abstract}
Domain adaptation (DA) tries to tackle the scenarios when the test data does not fully follow the same distribution of the training data, and multi-source domain adaptation (MSDA) is very attractive for real world applications. By learning from large-scale unlabeled samples, self-supervised learning has now become a new trend in deep learning. It is worth noting that both self-supervised learning and multi-source domain adaptation share a similar goal: they both aim to leverage unlabeled data to learn more expressive representations. Unfortunately, traditional multi-task self-supervised learning faces two challenges: (1) the pretext task may not strongly relate to the downstream task, thus it could be difficult to learn useful knowledge being shared from the pretext task to the target task; (2) when the same feature extractor is shared between the pretext task and the downstream one and only different prediction heads are used, it is ineffective to enable inter-task information exchange and knowledge sharing. To address these issues, we propose a novel \textbf{S}elf-\textbf{S}upervised \textbf{G}raph Neural Network (SSG), where a graph neural network is used as the bridge to enable more effective inter-task information exchange and knowledge sharing. More expressive representation is learned by adopting a mask token strategy to mask some domain information. Our extensive experiments have demonstrated that our proposed SSG method has achieved state-of-the-art results over four multi-source domain adaptation datasets, which have shown the effectiveness of our proposed SSG method from different aspects. Code is available at \textcolor{red}{https://github.com/a791702141/SSG}.

\end{abstract}



\keywords{multi-source domain adaptation, self-supervised learning, graph neural network}


\maketitle

\section{Introduction}
The success of deep learning largely depends on the assumption that the test data follows the same distribution as the training data. However, the collected data often do not fully conform to this assumption, which will cause a huge performance deterioration of deep models. To tackle this distribution shift issue, domain adaptation (DA) \cite{tzeng2014deep, sun2016deep, deng2019cluster} has been proposed  \cite{quinonero2008dataset}. DA often concentrates on transferring the knowledge learned from the labeled source domain to the unlabeled target domain, and multi-source domain adaptation (MSDA) is more valuable because we often have data from more than one domain in real world applications. 

\begin{figure}
	\centering
	\includegraphics[width=\columnwidth]{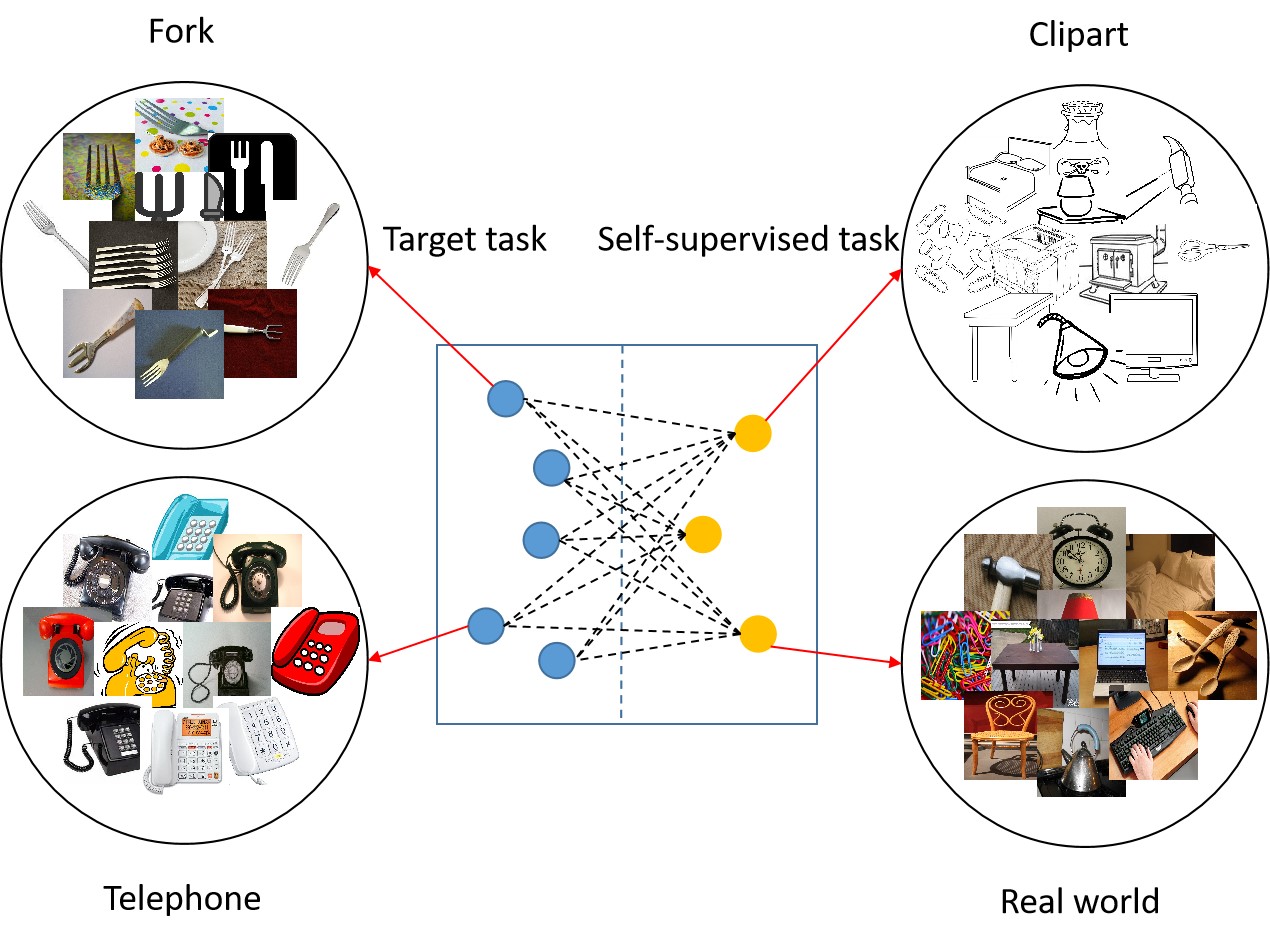}
	\caption{This figure shows the general framework of our proposed self-supervised graph neural network method. The blue nodes represent the embedding of categories in the target task, and the yellow nodes represent the embedding of domains in the self-supervised task. The graph neural network establishes the connection between the self-supervised task and the target task.}
	\label{fig1}
\end{figure}

By learning from large-scale unlabeled samples, self-supervised learning \cite{gidaris2019boosting, su2020does, lee2020self} has now become a new trend in deep learning. Self-supervised learning focuses on learning more expressive representations from large-scale unlabeled samples with/without few labeled samples for the pretext tasks. Multi-source domain adaptation aims to learn a generalized model by leveraging both the unlabeled data from the target domain and the labeled data from multiple source domains. In other words, MSDA can exploit large amounts of unlabeled data from the target domain to extend the size and the variety of the training data through self-supervised learning. It is worth noting that both self-supervised learning and multi-source domain adaptation share a similar goal: both of them aim to leverage large amounts of unlabeled data to learn more expressive representations. This observation motivates us to combine self-supervised learning with multi-source domain adaptation for learning more adequate representations.

In regular self-supervised learning, the purpose is to leverage the shared feature extractors (learned from the pretext tasks) for the target/downstream tasks, so their inter-task relevance is an issue that must be considered \cite{su2020does,you2020does}. To some extent, the stronger the correlation, the more meaningful the learned features for the target task. On the other hand, domain information is implicit knowledge like color, location and angle. Therefore, we consider the domain information as the labels of the pretext task instead of traditional self-supervised tasks, such as predicting the color \cite{larsson2016learning, zhang2016colorful}, angle \cite{komodakis2018unsupervised, feng2019self}, or position of image patches \cite{doersch2015unsupervised, noroozi2016unsupervised}. The deep model can benefit from the fact that both the domain information and the category labels are significantly correlated: for the same category, it could have different representations in different domains.

We can treat MSDA as a self-supervised learning issue with two tasks: the target task is category classification, while the pretext task is domain classification, thus MSDA can also be treated as a multi-task learning problem. Traditional multi-task self-supervised methods \cite{gidaris2019boosting,lee2020self, su2020does} often adopted a shared feature extractor and different linear prediction heads for the target task and the self-supervised one. However, these approaches require the feature extractor to learn the commonly-shared knowledge for the pretext task and the target task simultaneously, which is ineffective. Ideally, we expect the pretext task to exchange the information and share the knowledge directly with the target task, so that the target task can gain valuable knowledge from the pretext task. Inspired by the remarkable ability of the graph for information exchange, we employ a graph neural network to integrate these two tasks by representing each category and each domain as a node embedding in a graph instead of using a different linear prediction head. The main idea is shown in Figure~\ref{fig1}. 

In addition, we also exploit mask token strategy \cite{devlin2018bert, lanchantin2021general} to further strengthen the learned representation in the auxiliary task. During the training phase, we randomly set the state of domain information as either positive or negative instead of masking them out as unknown. Since a graph neural network is employed in our method, this masking strategy is realized by adding perturbations to the features for the graph nodes.

Overall, our contributions are as follows:

\begin{itemize}
    \item We propose a novel \textbf{S}elf-\textbf{S}upervised \textbf{G}raph Neural Network (SSG) to enable multi-source domain adaptation, where domain classification is adopted as the pretext task and the graph neural network is used as the bridge between the self-supervised task and the target task.
    \item We use a mask token scheme to make the training phrase stable and enhance domain representation learning.
    \item Our experiments have demonstrated that our proposed SSG method has achieved state-of-the-art results on four multi-source domain adaptation datasets, and other experiments have also verified the effectiveness of our proposed SSG method from different aspects.
\end{itemize}

\section{Related Works}
\subsection{Unsupervised Domain Adaptation}
Unsupervised domain adaptation (UDA) \cite{wilson2020survey, jiang2020implicit} aims to solve the issue of leveraging unlabeled data from the target domain to generalize the model learned from the labeled data for the source domain to the target domain. The mainstream methods usually tried to learn a domain-invariant feature representation, where the features for the same category follow a similar distribution no matter what domain it comes from \cite{baktashmotlagh2013unsupervised, zhao2019learning}. Some methods \cite{gretton2006kernel, gretton2012kernel, tzeng2014deep, sun2016deep, lee2019sliced, kang2019contrastive} solved this problem by minimizing the domain discrepancy, including maximum mean discrepancy (MMD), correlation alignment (CORAL) and contrastive domain discrepancy (CCD). Another important branch of UDA exploited adversarial network \cite{ghifary2016deep, bousmalis2017unsupervised, zhang2018collaborative, long2018conditional} to align the source domain with the target domain at a feature or pixel level by the adversarial networks or generative adversarial networks (GAN). 

\subsection{Multi-Source Domain Adaptation}
Instead of studying domain adaptation from one single source domain, multi-source domain adaptation become a more realistic problem \cite{mansour2008domain, crammer2008learning, ben2010theory, mancini2018boosting, xu2018deep}. Zhao et al. \cite{zhao2018adversarial} exploited an adversarial network to solve this problem. Peng et al. \cite{peng2019moment} dynamically aligned moments of multi-source domain feature distributions. Nguyen et al. \cite{nguyen2021most} tried to reduce the gap between the mixture of the source domain and the target domain by adopting student-teacher learning. Zhou et al. \cite{zhou2021domain} trained each source domain as an expert and let these experts leverage complementary information to get a more generalized model in the target domain. In general, prevailing methods mainly focused on adversarial learning and distillation methods. 

\begin{figure*}
	\centering
	\includegraphics[width=\textwidth]{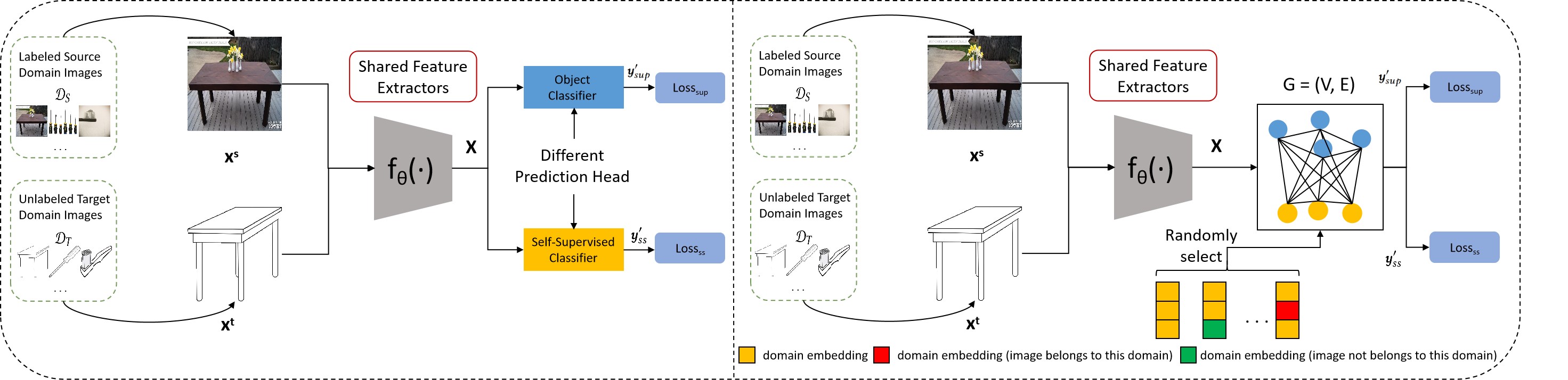}
	\caption{This figure shows the overall framework of our proposed method. The left half exhibits the traditional multi-task self-supervised learning framework by adopting a shared feature extractor and different heads. The right half displays our proposed method by using a graph neural network to exchange information between the pretext task and the target task. The embedding selection in the lower-right shows the mask domain process. We will randomly select a representation of domain embeddings for the graph neural network. The red one represents that this image belongs to this domain, while the green one indicates that this image does not belong to this domain.}
	\label{fig2}
\end{figure*}

\subsection{Self-Supervised Learning}
Self-supervised learning provides robust representation by transferring knowledge from the pretext tasks without extra annotations \cite{hendrycks2019using,gidaris2019boosting,lee2020self}. The core of leveraging self-supervision lies in two aspects. The first one is how to select the pretext tasks, and the conventional solution is to predict colors \cite{larsson2016learning, zhang2016colorful}, angle \cite{komodakis2018unsupervised, feng2019self} of images or the locations of image patches \cite{doersch2015unsupervised, noroozi2016unsupervised}. The other problem is how to bridge the self-supervised task and the main task, and the solutions mainly fall under two parts: pre-training $\&$ fine-tuning or multi-task learning. In the pre-training $\&$ fine-tuning scheme, the model is firstly pre-trained in the self-supervised task and then fine-tuned in the target task \cite{noroozi2016unsupervised, komodakis2018unsupervised, trinh2019selfie}. In the multi-task learning scheme, a joint objective is used to bridge the self-supervised task with the target supervised task \cite{doersch2017multi, ren2018cross, gidaris2019boosting}. Taking these two core issues into consideration, we adopt the domain classification as the pretext task for supervised category classification and exploit graph neural networks to bridge the connections between the self-supervised task and the target supervised task in a multi-task learning manner.

\section{Method}

\subsection{Preliminaries}

In regular self-supervised learning, models are often pre-trained in a self-supervised task and fine-tuned in the target task \cite{you2020does}. In the pre-training stage, the self-supervised task ("ss") is trained as follows:

\begin{equation}
\begin{aligned}
\boldsymbol{y^{\prime}}_{\mathrm{ss}} &=f_{\theta}(\boldsymbol{X}_{\mathrm{ss}})  , \\
\theta_{\mathrm{ss}}^{*} &=\arg \min _{\theta} \mathcal{L}_{\mathrm{ss}}(\theta) \\
&=\arg \min _{\theta} L_{\mathrm{ss}}\left(\boldsymbol{y^{\prime}}_{\mathrm{ss}}, \boldsymbol{y}_{\mathrm{ss}}\right).
\end{aligned}
\end{equation}

During fine-tuning stage, the model is fine-tuned in the target task using parameter $ \theta $, which is initialized by $\theta_{ss}^{*}$. 

Another scheme of self-supervision is multi-task learning, where self-supervised tasks are acted as auxiliary regularizations \cite{gidaris2019boosting, lee2020self, you2020does}. This process is formulated as: 

\begin{equation}
\begin{aligned}
\boldsymbol{y^{\prime}} &=f_{\theta}(\boldsymbol{X})  , \quad \boldsymbol{y^{\prime}}_{\mathrm{ss}}=f_{\theta}(\boldsymbol{X}_{\mathrm{ss}}) ,  \\
\theta^{*} &=\arg \min _{\theta} \alpha_{1} \mathcal{L}_{\mathrm{sup}}(\theta)+\alpha_{2} \mathcal{L}_{\mathrm{ss}}(\theta),
\end{aligned}
\end{equation}
where $ \mathcal{L}_{\mathrm{sup}} $ is the loss function of target supervised task and $\mathcal{L}_{\mathrm{ss}}$ is the loss function of pretext self-supervised task.

\subsection{Self-Supervised Multi-Source Domain Adaptation}

For multi-source domain adaptation with $ n $ domains, we have $n-1$ source domains dataset $\mathcal{D}_{S} = \{\mathcal{D}_{1},\dots, \mathcal{D}_{n-1}\} $ to learn a model with great generalization performance in the target domain $ \mathcal{D}_{T} $. Among them, the source domain owns the data and the label, while the target domain has only the unlabeled data. Different from single-source domain adaptation, multi-source domain adaptation usually has several source domains, so $ n $ is always larger than 2. 

For the input images and labels, we define them as $ \boldsymbol{x} $ and $ \boldsymbol{y} $, while the images from source domains are $ \boldsymbol{x}^{s} $ and the images from target domain are $ \boldsymbol{x}^{t} $. These images are first fed into a convolutional neural network (CNN) $ f $ to extract features and get corresponding $ \boldsymbol{X} \in \mathbb{R}^{D} $ while $D$ is the feature dimension.


For a MSDA task with $ n $ domains and $ c $ categories, we build a graph $ G = (V, E) $ where we model each domain and each category as a node of the graph $V = \{v_{1}, \dots , v_{n+c}\}$ and exploit a d-dimensional node embedding $ \boldsymbol{Z} \in \mathbb{R}^{(n+c) \times d} $ to represent the nodes.

\begin{figure*}[h]
	\centering
	\includegraphics[width=\textwidth]{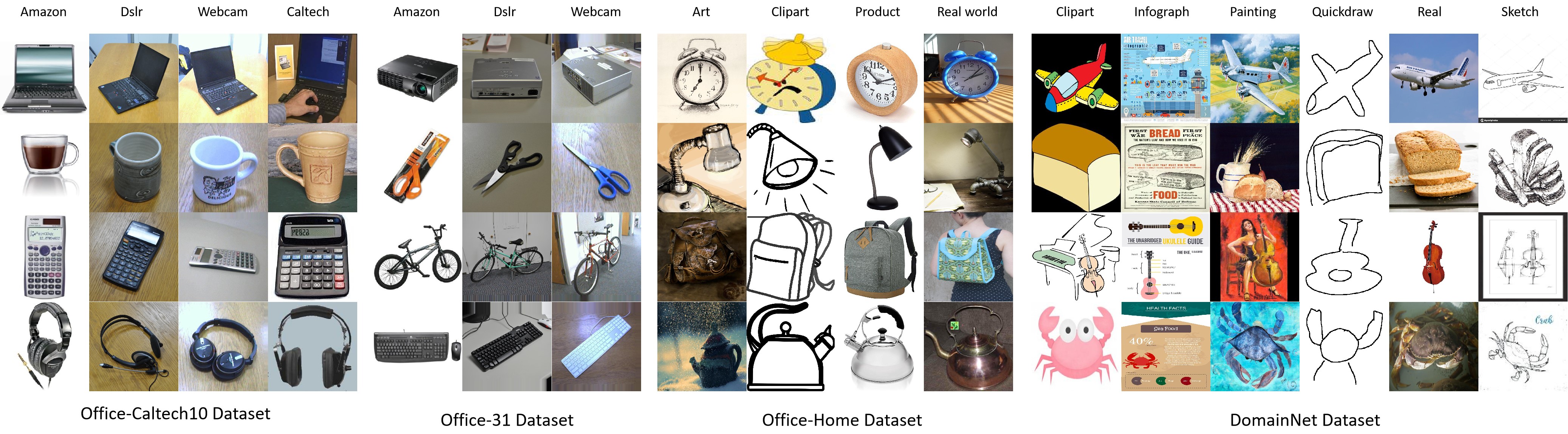}
	\caption{This figure shows the dataset we adopted for the experiments. The upper-left is the Office-Caltech10 dataset. The lower-left is the Office-31 dataset. The upper-right is the Office-Home dataset. The lower-right is the DomainNet dataset.}
	\label{fig3}
\end{figure*}

\begin{table*}[h]
	\caption{Comparison with state-of-the-art models on Office-Caltech10 dataset}
	\label{Table1}
	\centering
	\begin{tabular}{ccccccc}
	\toprule
	Standards &     Methods    & $\rightarrow$ D & $\rightarrow$ W & $\rightarrow$ A & $\rightarrow$ C & Avg \\
	\midrule
		\multirow{2}{*}{Source Combine}
		& Source-only & 98.3 &  99.0  & 86.1 & 87.8  & 92.8 \\
		& DAN \cite{long2015learning} & 98.2 &  99.3 & 94.8 & 89.7 & 95.5 \\
		\midrule
		\multirow{8}{*}{Multi-Source}  
        & DAN \cite{long2015learning} & 99.1 & 99.5 & 91.6 & 89.2 & 94.8 \\
        & JAN \cite{long2017deep} & 99.4 & 99.4 & 91.8 & 91.2 & 95.5 \\
        & DCTN \cite{xu2018deep} & 99.0 & 99.4 & 92.7 & 90.2 & 95.3 \\
        & MCD \cite{saito2018maximum} & 99.1 & 99.5 & 92.1 & 91.5 & 95.6 \\
        & $\text{M}^3$SDA \cite{peng2019moment} & 99.2 & 99.5 & 94.5 & 92.2 & 96.4 \\ 
        & CMSS \cite{yang2020curriculum} & 99.3 & 99.6 & 96.6 & 93.7 & 97.2 \\
		& STEM \cite{nguyen2021stem} & \bf{100} & \bf{100} & 98.4 & \bf{94.2} & 98.2 \\
		& SSG \emph{(ours)} & \bf{100} & \bf{100}  & \bf{99.0} & \bf{94.2} & \bf{98.3} \\
		
    \bottomrule
	\end{tabular}
\end{table*}

To model the relationship of each node, we use Gaussian kernel $\mathcal{K}_{G}$ to generate adjacency matrix $\boldsymbol{A}$ of the graph $ G $:

\begin{equation}
\boldsymbol{A}_{i,j}=\mathcal{K}_{G}\left(\boldsymbol{Z}_{i}^{\mathrm{T}},\boldsymbol{Z}_{j}^{\mathrm{T}}\right)=\exp \left(-\frac{\left\|\boldsymbol{Z}_{i}^{\mathrm{T}}-\boldsymbol{Z}_{j}^{\mathrm{T}}\right\|_{2}^{2}}{2 \sigma^{2}}\right),
\end{equation}
where $ \sigma $ is a hyperparameter that is empirically set as 0.005. 

With the node embedding $ \boldsymbol{Z} $ and the adjacency matrix $ \boldsymbol{A} $, we adopt graph convolutional network (GCN) \cite{estrach2014spectral, kipf2016semi} to aggregate the information of neighbor nodes: 

\begin{equation}
	\boldsymbol{H}^{(l+1)}=\sigma\left(\boldsymbol{\tilde{D}}^{-\frac{1}{2}} \boldsymbol{\tilde{A}} \boldsymbol{\tilde{D}}^{-\frac{1}{2}} \boldsymbol{H}^{(l)} \boldsymbol{W}^{(l)}\right),
\end{equation}
where $\boldsymbol{\tilde{A}} \in \mathbb{R}^{(n+c) \times (n+c)}$ is the adjacency matrix $ \boldsymbol{A} $ with self-connecting edges, $ \boldsymbol{\tilde{D}} $ is the degree matrix of $\boldsymbol{\tilde{A}}$ and $ \boldsymbol{W}^{(l)} $ is a learnable weight matrix. Node embedding $ \boldsymbol{Z} $ is inputted as $ \boldsymbol{H}^{0} $ in the first layer and the output of the last layer is the extracted graph feature $ \boldsymbol{Z}^{\prime} \in \mathbb{R}^{(n+c) \times D} $, where the dimension $ D $ is same to the extracted image feature $ \boldsymbol{X} $. 

Subsequently, the prediction $ \boldsymbol{y}^{\prime} $ is calculated as: 

\begin{equation}
    \boldsymbol{y}^{\prime} = \boldsymbol{Z}^{\prime}\boldsymbol{X} = \boldsymbol{Z}^{\prime}\sigma(f_{\theta_{CNN}}(\boldsymbol{x})), \label{head}
\end{equation}
where $ \sigma(\cdot) $ is the activation function.

Given the prediction $ \boldsymbol{y}^{\prime} $, we divide it into self-supervised domain prediction $ \boldsymbol{y}_{ss}^{\prime} $ and the supervised category prediction $ \boldsymbol{y}_{sup}^{\prime} $. Then we use a multi-task learning manner to optimize the parameters: 

\begin{equation}
\mathcal{L}_{MT} = \alpha_{1}\mathcal{L}_{\mathrm{sup}}(\boldsymbol{y}_{sup}^{\prime}, \boldsymbol{y}_{sup})+\alpha_{2} \mathcal{L}_{\mathrm{ss}}(\boldsymbol{y}_{ss}^{\prime}, \boldsymbol{y}_{ss}), \label{alpha}
\end{equation}
where $ \boldsymbol{y}_{sup} $ is the category label and $ \boldsymbol{y}_{ss} $ is the domain label.

For $ \mathcal{L}_{\mathrm{sup}} $ term, it contains the source domain loss and the target domain loss:

\begin{equation}
    \mathcal{L}_{\mathrm{sup}} = \mathcal{L}_{src} + \lambda\mathcal{L}_{tgt} = \mathcal{L}_{ce}(\boldsymbol{y}_{src}^{\prime}, \boldsymbol{y}_{src}) + \lambda\left(-\sum \boldsymbol{y}_{tgt}^{\prime} \log \left(\boldsymbol{y}_{tgt}^{\prime}\right)\right), \label{lamda}
\end{equation}
where $ \mathcal{L}_{ce} $ represent the cross-entropy loss.

For $ \mathcal{L}_{\mathrm{ss}} $ term, we also use the cross-entropy loss $ \mathcal{L}_{ce} $ for domain classification:
\begin{equation}
    \mathcal{L}_{\mathrm{ss}} = \mathcal{L}_{ce}(\boldsymbol{y}_{dmn}^{\prime}, \boldsymbol{y}_{dmn}).
\end{equation}


For the self-supervised loss, we extend the unlabeled data from the target domain to the self-supervised task. Learning the target domain information is able to establish the connections between the source domain and the target domain, thereby enhancing the representation ability of the convolutional neural network and helping the category classification task learn more adequate knowledge of the target domain.

\subsection{Mask Token Strategy}

The mask token strategy is proposed in language tasks \cite{devlin2018bert} to enhance the ability of representations learned by the self-supervised task according to randomly masking and predicting some tokens. 

\begin{table*}[h]
	\caption{Comparison with state-of-the-art models on Office-31 dataset}
	\label{Table2}
	\centering
	\begin{tabular}{cccccc}
	\toprule
	Standards &     Methods    & $\rightarrow$ D & $\rightarrow$ W & $\rightarrow$ A & Avg \\
	\midrule
		\multirow{5}{*}{Single Best} & Source-only & 99.0 & 95.3 & 50.2 & 81.5   \\

        & RevGrad \cite{ganin2015unsupervised} & 99.2 & 96.4 & 53.4 & 83.0 \\
		& DAN \cite{long2015learning} & 99.0 & 96.0 & 54.0 & 83.0  \\
		& RTN \cite{long2016unsupervised} & 99.6 & 96.8 & 51.0 & 82.5 \\
		& ADDA \cite{tzeng2017adversarial} & 99.4 & 95.3 & 54.6 & 83.1  \\
		\midrule
		\multirow{5}{*}{Source Combine}& DAN \cite{long2015learning} & 98.8 & 96.2 & 54.9 & 83.3 \\
		& RTN \cite{long2016unsupervised} & 99.2 & 95.8 & 53.4 & 82.8  \\
		& JAN \cite{long2017deep} & 99.4 & 95.9 & 54.6 & 83.3 \\
		& ADDA \cite{tzeng2017adversarial}  & 99.2 & 96.0 & 55.9 & 83.7 \\
		& MCD \cite{saito2018maximum} & 99.5 & 96.2 & 54.4 & 83.4 \\
		\midrule
		\multirow{6}{*}{Multi-Source}  & MDAN \cite{zhao2018adversarial} & 99.2 & 95.4 & 55.2 & 83.3  \\
		& DCTN \cite{xu2018deep} & 99.6 & 96.9 & 54.9 & 83.8 \\
		& MDDA \cite{zhao2020multi} & 99.2 & 97.1 & 56.2 & 84.2 \\
		& LtC-MSDA \cite{wang2020learning} & 99.6 & 97.2 & 56.9 & 84.6 \\
		& MOST \cite{nguyen2021most} & \bf{100} & 98.7 & 60.6 & 86.4 \\
		& SSG \emph{(ours)} & \bf{100} & \bf{99.5} & \bf{71.3} & \bf{90.3} \\
		
    \bottomrule
	\end{tabular}
\end{table*}

\begin{table*}[h]
	\caption{Comparison with state-of-the-art models on Office-Home dataset}
	\label{Table3}
	\centering
	\begin{tabular}{cccccccc}
	\toprule
	Standards &     Methods    & $\rightarrow$ A & $\rightarrow$ C & $\rightarrow$ P & $\rightarrow$ R & Avg \\
	\midrule
		\multirow{5}{*}{Single Best} & Source-only & 65.3 & 49.6 & 79.7 & 75.4 & 67.5   \\
		& DDC \cite{tzeng2014deep} & 64.1 & 50.8 & 78.2 & 75.0 & 67.0 \\
		& DAN \cite{long2015learning} & 68.2 & 56.5 & 80.3 & 75.9 & 70.2  \\
		& RevGrad \cite{ganin2015unsupervised} & 67.9 & 55.9 & 80.4 & 75.8 & 70.0 \\
		& CORAL \cite{sun2017correlation} & 67.0 & 53.6 & 80.3 & 76.3 & 69.3  \\
		\midrule
		\multirow{3}{*}{Source Combine}& DAN \cite{long2015learning} & 68.5 & 59.4 & 79.0 & 82.5 & 72.4 \\
		& RevGrad \cite{ganin2015unsupervised} & 68.4 & 59.1 & 79.5 & 82.7 & 72.4 \\
		& CORAL \cite{sun2017correlation} & 68.1 & 58.6 & 79.5 & 82.7 & 72.2  \\
		\midrule
		\multirow{4}{*}{Multi-Source}  
		& MFSAN \cite{zhu2019aligning} & 72.1 & 62.0 & 80.3 & 81.8 & 74.1  \\
		& SImpAl \cite{venkat2020your} & 70.8 & 56.3 & 80.2 & 81.5 & 72.2  \\
		& DARN \cite{wen2020domain} & 70.0 & \bf{68.4} & 82.8 & 83.9 &  76.3 \\
		& SSG \emph{(ours)} & \bf{81.7±0.40} & 66.7±0.61  & \bf{86.0±0.35} & \bf{84.9±0.27} & \bf{79.8} \\
		
    \bottomrule
	\end{tabular}
\end{table*}


We also adopt this technique to learn more robust representations in SSG by treating each domain as a token. For the self-supervised domain classification task, we randomly mask some domain information and ask our model to predict the domain instead of directly predicting all domains. The advantages of this technique are as follows: (1) Compared with directly predicting all domains, the training stage will be more stable with extra domain information. (2) Mask token gives different graph node representations for the same image, thereby expanding the sample representations in the graph.

The next question is how to give or hide the domain information to the model as a prior. We have adopted a graph neural network to integrate the domains and the categories, so our idea is to exploit a perturbation on graph node embedding to represent whether this image belongs to this domain or not. Since we already have the domain labels, we decide to use the predefined perturbations to represent this semantic information like \cite{lanchantin2021general}.

For the aforementioned node embedding $ \boldsymbol{Z} $, we use a predefined perturbation $ \boldsymbol{S} $ to represent whether this image belongs to this domain or not. In other words, if we randomly decide to give the domain information, the perturbed node representation will be obtained as: 

\begin{equation}
    \begin{aligned}
    \boldsymbol{Z}_{i} = \boldsymbol{Z}_{i} + \boldsymbol{S} \\
    \boldsymbol{Z}_{i} = \boldsymbol{Z}_{i} - \boldsymbol{S}, \label{state}
    \end{aligned}
\end{equation}
where increasing perturbation represents that this image belongs to the $i$-th domain and reducing the perturbation means that this image does not belong to the $i$-th domain. Note that the change of $ \boldsymbol{Z} $ is after the update of the adjacency matrix $ \boldsymbol{A} $. To learn this special representation pattern, the model is asked to predict the domain for each image, whether the domain information is given or not. Our whole framework is shown in Figure~\ref{fig2}. 


\section{Experiments}

\subsection{Datasets and Experimental Settings}

\begin{table*}[h]
	\caption{Comparison with state-of-the-art models on DomainNet dataset. For a fair comparison, $"*"$ reproduces the method in our experiment setting without the pre-trained features help of its original paper.}
	\label{Table4}
	\centering
	\begin{tabular}{cccccccccc}
	\toprule
	Standards &     Methods    & $\rightarrow$ C & $\rightarrow$ I & $\rightarrow$ P & $\rightarrow$ Q  & $\rightarrow$ R & $\rightarrow$ S &   Avg \\
	\midrule
		\multirow{8}{*}{Single Best} & Source Only & 39.6±0.58 & 8.2±0.75 & 33.9 ± 0.62 & 11.8±0.69 & 41.6±0.84 & 23.1±0.72 & 26.4   \\
		& DAN \cite{long2015learning} & 39.1±0.51 & 11.4±0.81 & 33.3±0.62 & 16.2±0.38 & 42.1±0.73 & 29.7±0.93 & 28.6  \\
		& DANN \cite{ganin2015unsupervised} & 37.9±0.69 & 11.4±0.91 & 33.9±0.60 & 13.7±0.56 & 41.5±0.67 & 28.6±0.63 & 27.8 \\
		& RTN \cite{long2016unsupervised} & 35.3±0.73 & 10.7±0.61 & 31.7±0.82 & 13.1±0.68 & 40.6±0.55 & 26.5±0.78 & 26.3  \\
		& JAN \cite{long2017deep} & 35.3±0.71 & 9.1±0.63 & 32.5±0.65 & 14.3±0.62 & 43.1±0.78 & 25.7±0.61 & 26.7  \\
		& ADDA \cite{tzeng2017adversarial} & 39.5±0.81 & 14.5±0.69 & 29.1±0.78 & 14.9±0.54 & 41.9±0.82 & 30.7±0.68 & 28.4 \\
		& SE \cite{french2017self} & 31.7±0.70 & 12.9±0.58 & 19.9±0.75 & 7.7±0.44 & 33.4±0.56 & 26.3±0.50 & 22.0 \\
		\midrule
		\multirow{8}{*}{Source Combine}
	& Source Only & 47.6±0.52 & 13.0±0.41 & 38.1±0.45 & 13.3±0.39 & 51.9±0.85 & 33.7±0.54 & 32.9 \\
    & DAN \cite{long2015learning} & 45.4±0.49 & 12.8±0.86 & 36.2±0.58 & 15.3±0.37 & 48.6±0.72 & 34.0±0.54 & 32.1 \\
    & DANN \cite{ganin2015unsupervised} & 45.5±0.59 & 13.1±0.72 & 37.0±0.69 & 13.2±0.77 & 48.9±0.65 & 31.8±0.62 & 32.6 \\
    & RTN \cite{long2016unsupervised} & 44.2±0.57 & 12.6±0.73 & 35.3±0.59 & 14.6±0.76 & 48.4±0.67 & 31.7±0.73 & 31.1 \\
    & JAN \cite{long2017deep} & 40.9±0.43 & 11.1±0.61 & 35.4±0.50 & 12.1±0.67 & 45.8±0.59 & 32.3±0.63 & 29.6 \\
    & ADDA \cite{tzeng2017adversarial} & 47.5±0.76 & 11.4±0.67 & 36.7±0.53 & 14.7±0.50 & 49.1±0.82 & 33.5±0.49 &32.2 \\
    & SE \cite{french2017self} & 24.7±0.32 & 3.9±0.47 & 12.7±0.35 & 7.1±0.46 & 22.8±0.51 & 9.1±0.49 & 16.1 \\
    & MCD \cite{saito2018maximum} & 54.3±0.64 & 22.1±0.70 & 45.7±0.63 & 7.6±0.49 & 58.4±0.65 & 43.5±0.57 & 38.5 \\
		\midrule
		\multirow{7}{*}{Multi-Source} 

		& DCTN \cite{xu2018deep} & 48.6±0.70 & 23.5±0.60 & 48.8±0.60 & 7.2±0.50 & 53.5±0.60 & 47.3±0.50 & 38.2 \\
		& MDAN \cite{zhao2018adversarial} & 52.4±0.60 & 21.3±0.80 & 46.9±0.40 & 8.6±0.60 & 54.9±0.60 & 46.5±0.70 & 38.4 \\
		& $ \text{M}^{3}\text{SDA} $ \cite{peng2019moment} & 58.6±0.53 & 26.0±0.89 & 52.3±0.55 & 6.3±0.58 & 62.7±0.51 & 49.5±0.76 & 42.6  \\
		& MDDA \cite{zhao2020multi} & 59.4±0.60 & 23.8±0.80 & 53.2±0.60 & 12.5±0.60 & 61.8±0.50 & 48.6±0.80 & 43.2 \\
		& CMSS \cite{yang2020curriculum} & 64.2±0.18 & 28.0±0.20 & 53.6±0.39 & 16.0±0.12 & 63.4±0.21 & 53.8±0.35 & 46.5  \\
        & LtC-MSDA \cite{wang2020learning} & 63.1±0.50 & \bf{28.7±0.70} & 56.1±0.50 & 16.3±0.50 & 66.1±0.60 & 53.8±0.60 & 47.4 \\
        & STE$\text{M}^{*} $ \cite{nguyen2021stem} & 38.9±0.17 & 11.7±0.11 & 29.4±0.47 & 8.4±0.07 & 37.2±0.22 & 33.3±0.31 & 26.5 \\
        & DAEL \cite{zhou2021domain} & \bf{70.8±0.14} & 26.5±0.13 & \bf{57.4±0.28} & 12.2±0.70 & 65.0±0.23 & \bf{60.6±0.25} & 48.7  \\
		& SSG \emph{(ours)} & 68.7±0.31 & 24.8±0.22 & 55.7±0.17 & \bf{18.4±0.06} & \bf{68.8±0.07} &  56.3±0.27 & \bf{48.8} \\
		
    \bottomrule
	\end{tabular}
\end{table*}

We use Office-Caltech10 \cite{gong2012geodesic}, Office-31 \cite{saenko2010adapting}, Office-Home \cite{venkateswara2017deep} and DomainNet \cite{peng2019moment} as our experimental benchmarks. Office-Caltech10 has 10 categories and 4 domains, including \emph{DSLR} (D), \emph{Webcam} (W), \emph{Amazon} (A) and \emph{Caltech} (C) with 2,533 images. Office-31 has 31 categories and 4,652 images, which belong to 3 domains: \emph{DSLR} (D), \emph{Webcam} (W) and \emph{Amazon} (A). Compared with Office-Caltech10, Office-31 has more categories and more images. Another dataset, called Office-Home, consists of 65 categories and 4 domains of \emph{Art} (A), \emph{Clipart} (C), \emph{Product} (P) and \emph{Real world} (R). This dataset contains 15,500 images with the common categories, including fork, table, etc. The recently proposed DomainNet is a larger dataset with 569,010 images of 345 categories in 6 domains: \emph{Clipart} (C), \emph{Infograph} (I), \emph{Painting} (P), \emph{Quickdraw} (Q), \emph{Real} (R) and \emph{Sketch} (S). As shown in Figure~\ref{fig3}, these four mainstream datasets can be divided two groups: the first group including Office-Caltech10 and Office-31 has a lower domain discrepancy and fewer categories; the second group including Office-Home and DomainNet holds a larger discrepancy (like \emph{Quickdraw} and \emph{Real} in DomainNet) and more categories.

For these datasets, we follow the experimental setting of \cite{peng2019moment, wang2020learning, nguyen2021stem, zhou2021domain}. stochastic gradient decent (SGD) \cite{sutskever2013importance} with learning rate of 1e-4 are applied to optimize the model, and ResNet-101 \cite{he2016deep} is pre-trained on ImageNet \cite{deng2009imagenet} as a feature extractor. $ \alpha_{1} $ and $ \alpha_{2} $ in Equation~\ref{alpha} are set as 1 and 0.1, respectively. The $ \boldsymbol{S} $ in Equation~\ref{state} are set as a D-dimension vector of 0.1. The $\lambda $ are empirically set as 5 in Equation~\ref{lamda}. The mask token ratio are set as 95\%, so 5\% prior domain knowledge is given during training. 

\subsection{Comparisons with the State-of-the-arts}

\begin{figure*}[h]
	\centering
	\includegraphics[width=\textwidth]{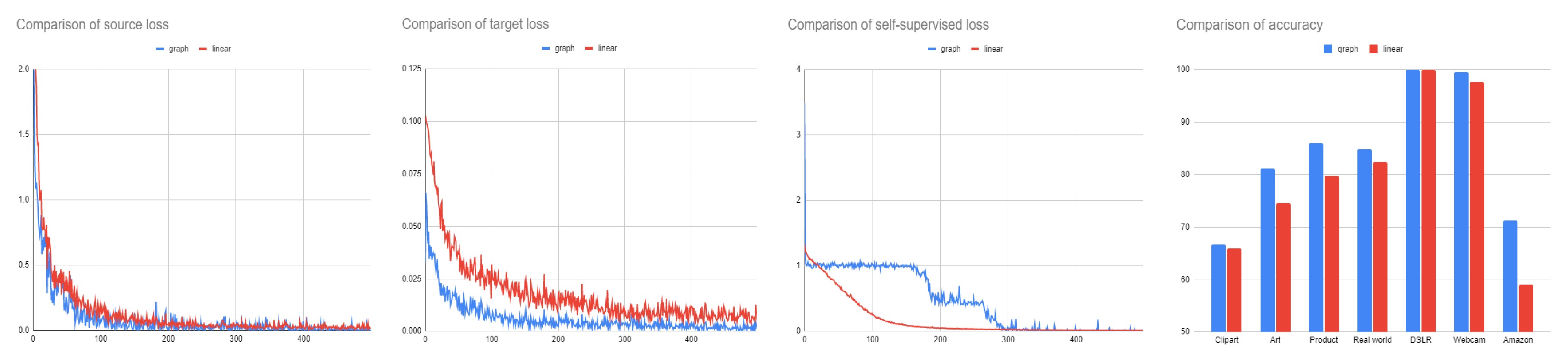}
	\caption{We use blue to represent our graph-based method and red to represent the linear-based method. Three line graphs on the left represent the trends of different loss functions, which is conducted in Office-31 with the target domain \emph{Webcam} of 500 epochs. The histogram on the right represents the overall accuracy of the two methods on Office-31 and Office-Home.}
	\label{fig4}
\end{figure*}

Generally, compared methods are split into three groups:
\begin{itemize}
\item \textbf{Single Best}: the works with the best single source domain adaptation results.
\item \textbf{Source Combine}: the methods consider multiple source domains as one domain for the domain adaptation.
\item \textbf{Multi-Source}: the methods adopt multiple source domains for the domain adaptation with the domain information for each image. 
\end{itemize}

For the Office-Caltech10 dataset, SSG achieves a state-of-the-art performance on the average of 98.3\% (shown in Table~\ref{Table1}). Due to the small number of categories in this dataset, most methods achieve high accuracy. Compared with the most competitive approach STEM \cite{nguyen2021stem}, our method can get the same 100\% in \emph{Webcam} and \emph{DSLR}, 94.2\% in \emph{Caltech}, and 0.6\% improved performance in \emph{Amazon}. 

For Office-31, SSG holds a clear advantage of 90.3\% accuracy (shown in Table~\ref{Table2}). Due to the small domain discrepancy, compared methods can get relatively high value in \emph{DSLR} and \emph{Webcam}, and SSG can get 99.5\% in \emph{Webcam}, outperforming the second-best MOST \cite{nguyen2021most} 0.8\%. Besides, our method achieves 71.3\% in \emph{Amazon}, with a large margin over other methods. On average, SSG increases the performance by 2.9\%. Another interesting observation is the performance difference caused by different standards. The state-of-the-art methods in the Source Combine group are better than the Single Best group, while the best model in the Multi-Source group is also better than the Source Combine group. Even for the same methods, like DAN \cite{long2015learning} and ADDA \cite{tzeng2017adversarial}, their Source Combine results are better than the Single Best results, and this observation continues to be found in the following two dataset results. We consider the underlying reason: (1) Learning knowledge from multiple domains is better than learning knowledge from a single domain, although performance gains may come from more examples. (2) Domain knowledge can help us promote the results by learning the domain discrepancy. 

As shown in Table~\ref{Table3}, Office-Home is a more challenging dataset with larger domain discrepancy, which makes methods hard to get high accuracy. In this dataset, SSG achieves 79.8\% accuracy with clear margins against all competitors. For \emph{Art}, SSG gets 81.7\% on average, which has an improvement of 9.6\% compared with the second-best method MFSAN \cite{zhu2019aligning}. For \emph{Clipart}, although SSG is not the best result, it is still the second-best result of 66.7\%, only 1.7\% difference with DARN \cite{wen2020domain}. For \emph{Product} and \emph{Real world}, our method also achieves the best performance of 86.0\% and 84.9\% with improvements of 3.2\% and 1.0\% over the second-best model. On average, our method beats DARN \cite{wen2020domain} 3.4\%. 

DomainNet is a larger dataset with more categories and domains. We compare our method with other works in Table~\ref{Table4}. In this dataset, a clear gap exists in Sing Best, Source Combine and Multi-source, further illustrating the importance of researching multi-source domain adaptation. For multi-source based methods, DCTN \cite{xu2018deep}, MDAN \cite{zhao2018adversarial} and CMSS \cite{yang2020curriculum} are adversarial learning based methods, which follow the conventional adversarial domain adaptation of a single source. These three methods can perform better than Single Best and Source Combine methods. Besides, MDDA \cite{zhao2020multi}, STEM \cite{nguyen2021stem} and DAEL \cite{zhou2021domain} are distillation based methods (without the pre-trained features help of the original paper, we reproduce STEM with the same extracted feature in our experiment), which concentrated on distilling the knowledge from source domains to target domain and achieved very competitive results. DAEL reached state-of-the-art results of 70.8\% in \emph{Clipart}, 57.4\% in \emph{Painting} and 60.6\% in \emph{Sketch}. Furthermore, $ \text{M}^{3}\text{SDA} $ \cite{peng2019moment} transferred knowledge by dynamically aligning moments of their feature distributions and proposed DomainNet dataset. LtC-MSDA \cite{wang2020learning} also used graph network and got state-of-the-art result of 28.7\% in \emph{Infograph}. Different from our method, they built a large graph of the domain number multiplied by the class number nodes and proposed relation alignment loss to connect graph parameters with feature extractors, whose overhead becomes high when the number of categories and domains increases. Compared with those methods, SSG achieves the state-of-the-art performance of 18.4\% in \emph{Quickdraw} and 68.8\% in \emph{Real}. For \emph{Clipart} and \emph{Sketch}, SSG achieves the second-best results after DAEL. On average, SSG reaches a state-of-the-art performance of 48.8\%.

\subsection{Self-Supervised Learning Study}

In conventional multi-task self-supervised learning, the same feature extractor and different linear prediction heads are often used as a scheme to connect the self-supervised task and the target task (shown in Figure~\ref{fig2}). These methods \cite{gidaris2019boosting,lee2020self, su2020does} often use linear layer in Equation~\ref{head}:

\begin{equation}
\begin{aligned}
    \boldsymbol{y}^{\prime}_{sup} = \boldsymbol{W}_{sup}\boldsymbol{X} = \boldsymbol{W}_{sup}\sigma(f_{\theta_{CNN}}(\boldsymbol{x})), \\
    \boldsymbol{y}^{\prime}_{ss} = \boldsymbol{W}_{ss}\boldsymbol{X} = \boldsymbol{W}_{ss}\sigma(f_{\theta_{CNN}}(\boldsymbol{x})), \label{head}
\end{aligned}
\end{equation}
where $ \boldsymbol{W}_{sup} $ and $ \boldsymbol{W}_{ss} $ is the linear mapping parameter in different prediction heads. Therefore, we compare this method with our proposed graph neural network based method.

As shown in Figure~\ref{fig4}, our proposed graph-based method gets a clear advantage on all target domains. Especially on \emph{Clipart} and \emph{Amazon}, graph-based method can surpass linear-based method 6.5\% and 12.3\%. These results verify the effectiveness of the graph-based self-supervised learning method. Subsequently, we try to find the underlying reason by plotting the line graph. We observe that the graph-based method converges faster, and the source and target loss value is continuously smaller than the linear-based method. However, the graph-based method shows a step-down trend for self-supervised loss, while the linear-based method shows a linear decline. Eventually, both of them converge to lower values. Interestingly, it seems that the graph-based method optimizes the source loss and the target loss in the early stage and only starts optimizing the self-supervised loss after the two converge. In the final stage, the graph-based method can converge to a lower loss value compared with the linear-based one, which may benefit from the information exchange between the self-supervised task and the target task in the graph network.

\subsection{Mask Ratio Study}

The ratio of masks is an issue worth investigating. Unlike masking a small proportion of tokens in linguistics task \cite{devlin2018bert}, tasks in the visual field have extracted features, so excessive domain information will make the model difficult to learn an appropriate representation. Therefore, we decide the mask ratio by experiments.

\begin{figure}[h]
	\centering
	\includegraphics[width=\columnwidth]{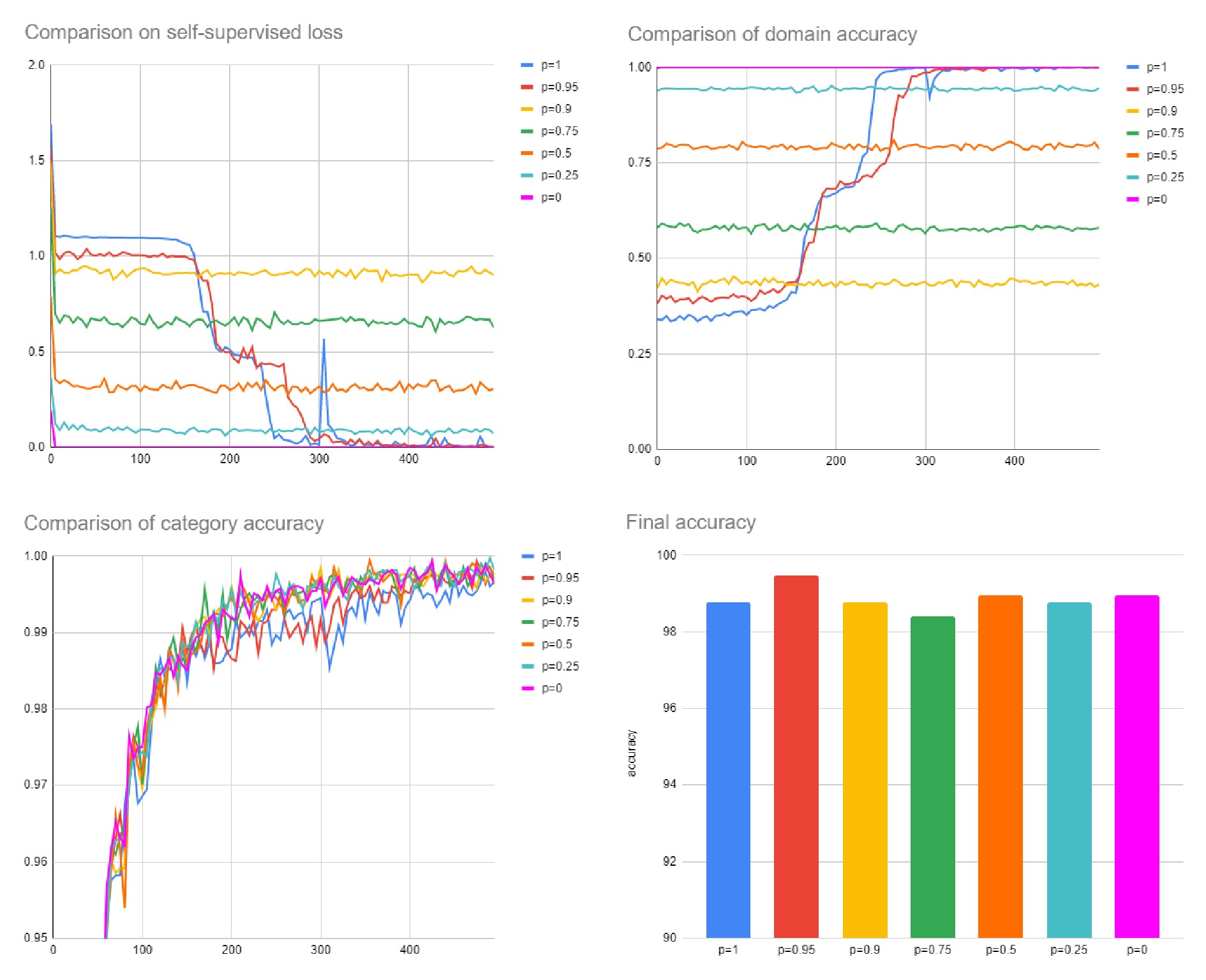}
	\caption{We study mask ratio and draw a line graph. The upper-left is the change of self-supervised loss and the upper-right is the domain accuracy in training stage. The lower-left is the category accuracy during training stage and the lower-right is the final accuracy in test phrase. All experiments are conducted in Office-31 with the target domain \emph{Webcam} of 500 epochs.}
	\label{fig5}
\end{figure}

As shown in Figure~\ref{fig5}, when 0\% domain information is masked, self-supervised loss quickly drops to 0 and domain accuracy quickly reaches 100\%. This reveals that the model can learn the domain information by adding perturbation. With the increase of mask ratio, domain loss falls rapidly and then starts to shock until the ratio reaches 95\%. For the mask ratio of 25\%, 50\%, 75\% and 90\%, excessive domain information causes the model to lose its ability to learn the proper knowledge. When 100\% domain is masked, it is theoretically equivalent to not using the mask token strategy. In this setting, the domain loss gradually decreases, and the training accuracy finally reaches 100\%, which illustrates that the domain is distinguishable even in the dataset of small domain discrepancy. However, the entire mask results are unstable and inferior to the 95\% mask results in the training and inference stages, so we ultimately determine the 95\% as our mask ratio.

\subsection{Ablation Study}

In addition to our core contribution of using the graph neural network for self-supervised learning, we explore the effectiveness of our other modules. Firstly, We conduct the ablation study for the loss function. Four versions are selected in this part including source loss only ($ \mathcal{L}_{src} $), source loss with target loss ($ \mathcal{L}_{src} + \mathcal{L}_{tgt} $), source loss with self-supervised domain loss ($ \mathcal{L}_{src} + \mathcal{L}_{ss} $) and all of these three losses (SSG). 

Besides, we also explore the influence of other modules of our proposed model. In this paper, we use trainable parameters as the node embeddings of the graph. The other choice is using the prototype-based method \cite{wang2020learning} to generate these node embeddings:

\begin{equation}
\boldsymbol{Z}_{k}=\frac{1}{\left|\mathcal{D}_{S}^{k}\right|} \sum_{\left(\boldsymbol{x}_{i}, \boldsymbol{y}_{i}\right) \in \mathcal{D}_{S}^{k}} f_{\theta_{CNN}}\left(\boldsymbol{x}_{i}\right),
\end{equation}
where $ \mathcal{D}_{S}^{k} $ represent the $k$-th category of source domain datasets $ \mathcal{D}_{S} $. Subsequently, we want to explore the influence of the mask token strategy. In general, we include two versions: the first is the SSG prototype, which uses prototype-based node embeddings. Another version does not adopt the mask token strategy and is named SSG w/o mask tokens.

As shown in Table~\ref{Table5} and Table~\ref{Table6}, experiments are conducted on Office-31 and Office-Home datasets. For the different loss function, the version that only uses source loss have an apparent decline compared to the final SSG, with 1.4\% and 4.0\% on average for Office-31 and Office-Home. Interestingly, the overall performance drops slightly when adding the self-supervised loss with source loss. The underlying reasons may be attributed to the over-learned relationship between the source domains, which weakens the generalization ability of the model in the target domain. The version source loss with target loss has a better performance than the source loss version, which relies on the relational feature extracted by the graph network. Moreover, the final version holds improvements compared with source loss and target loss results, illustrating that the combination of self-supervised loss and target loss can learn more helpful target domain knowledge.

For the node embeddings and mask token ablation study, when the prototype-based method is exploited, we observe that the performance deteriorates at 1.7\% and 1.1\% on average, respectively. These results show that the parameter-based node embedding is more expressive than the average-based prototype node embedding. Another try is not using the mask token strategy, which does not give the model any additional information, but merely connects the self-supervised and target tasks through a graph network. Compared with the final version, SSG w/o mask tokens has a 1.5\% and 1.1\% decline on two datasets, indicating that using parameter-based node embeddings is not the sole contributor to our approach.

In general, node embeddings and mask token strategy are not as sensitive as the loss combination, but the experimental results illustrate that all modules are essential for our proposed method.

\begin{table}[h]
	\caption{Ablation study on Office-31 dataset}
	\label{Table5}
	\centering
	\begin{tabular}{cccccc}
	\toprule
	   & Methods    & $\rightarrow$ D & $\rightarrow$ W & $\rightarrow$ A & Avg \\
	\midrule
	    & $ \mathcal{L}_{src} $ & 100 & 98.6 & 68.1 & 88.9   \\
	    & $ \mathcal{L}_{src} + \mathcal{L}_{tgt} $ & 98.2 & 99.5 & 71.1 & 89.6 \\
	    & $ \mathcal{L}_{src} + \mathcal{L}_{ss} $ & 99.1 & 97.8 & 67.0 & 88.0 \\
	    & SSG prototype & 99.1 & 98.9 & 67.7 & 88.6 \\
		& SSG w/o mask tokens & 99.1 & 99.5 & 67.7 & 88.8 \\
		& SSG ($ \mathcal{L}_{src} + \mathcal{L}_{tgt} + \mathcal{L}_{ss} $) & 100 & 99.5 & 71.3 & 90.3 \\
		
    \bottomrule
	\end{tabular}
\end{table}

\begin{table}[h]
	\caption{Ablation study on Office-Home dataset}
	\label{Table6}
	\centering
	\begin{tabular}{ccccccc}
	\toprule
	   & Methods    & $\rightarrow$ A & $\rightarrow$ C & $\rightarrow$ P & $\rightarrow$ R & Avg \\
	\midrule
	    & $ \mathcal{L}_{src} $ & 76.3 & 59.5 & 83.8 & 83.4 & 75.8 \\
	    & $ \mathcal{L}_{src} + \mathcal{L}_{tgt} $ & 79.9 & 65.0 & 84.9 & 84.1 & 78.5 \\
	    & $ \mathcal{L}_{src} + \mathcal{L}_{ss} $ & 76.1 & 59.5 & 84.0 & 83.2 & 75.7 \\
	    & SSG prototype & 80.5 & 67.2 & 83.8 & 83.2 & 78.7 \\
		& SSG w/o mask tokens & 79.5 & 65.7 & 85.5 & 84.1 & 78.7   \\
		& SSG ($ \mathcal{L}_{src} + \mathcal{L}_{tgt} + \mathcal{L}_{ss} $) & 81.7 & 66.7  & 86.0 & 84.9 & 79.8 \\
		
    \bottomrule
	\end{tabular}
\end{table}

\section{Conclusion}
This paper proposes a self-supervised graph neural network to achieve multi-source domain adaptation. The domain classification task is treated as the pretext task for the target category classification, and the graph neural network is used as the bridge between those two tasks by self-supervised learning. As a typical self-supervised framework, multi-task learning is used in the proposed model by attaching the auxiliary domain loss function. We have conducted extensive experiments, including comparison experiments with state-of-the-art methods and the confirmatory experiments of each module. Our experimental results have demonstrated that our proposed SSG method has achieved state-of-the-art results over four datasets.


\end{document}